\documentclass[11pt]{article}

\usepackage{acl}

\usepackage{times}
\usepackage{latexsym}
\usepackage[T1]{fontenc}
\usepackage[utf8]{inputenc}
\usepackage{microtype}
\usepackage{inconsolata}
\usepackage{graphicx}
\usepackage{amsmath}
\usepackage{amssymb}
\usepackage{booktabs}
\usepackage{tabularx}
\usepackage{multirow}
\usepackage{fancyhdr}
\usepackage{xcolor}
\usepackage{tcolorbox}
\usepackage{enumitem}

\title{Jupiter-N Technical Report}

\author{George Drayson \\
  Locai Labs, University College London \\
  \texttt{george.drayson@locailabs.com} \\}

\begin{document}
\thispagestyle{fancy}
\fancyhf{}
\fancyhead[C]{\includegraphics[height=0.9cm]{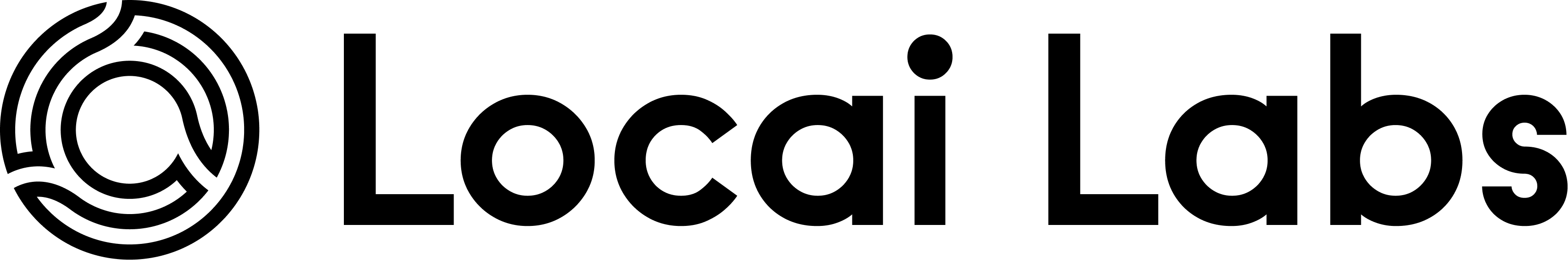}}
\renewcommand{\headrulewidth}{0pt}
\setlength{\headsep}{1.2cm}
\maketitle
\begin{abstract}
We present Jupiter-N, a hybrid reasoning model post-trained from Nemotron 3 Super, a fully open-source $120$ billion parameter LLM. We target three objectives: (1) agentic capability via uncertainty-curated trajectories; (2) UK cultural alignment via synthetic data grounded in cultural norms; and (3) Welsh language support via parallel corpora and LLM-translated Welsh conversations. Our data curation strategy carefully preserves the base model's capabilities: using our Forget-Me-Not framework, we mix on-policy synthetic replay with off-policy task data to mitigate catastrophic forgetting, and include a mixture of reasoning and non-reasoning traces to maintain Nemotron's hybrid reasoning ability. Jupiter-N achieves standout gains over Nemotron in Welsh ($+18$ on ARC-Easy, $+5.25$ on MMLU-Lite), terminal-use ($+9.1$ on Terminal Bench 2) and instruction following ($+4.4$ on IFBench), while retaining the base model capabilities. We frame this work as a reproducible template for \emph{sovereign post-training}: substituting cultural knowledge, institutional corpora, and target languages produces an equivalent pipeline for any country. All model weights\footnote{\href{https://huggingface.co/locailabs/Jupiter-N-120B}{\nolinkurl{locailabs/Jupiter-N-120B}}} and all post-training datasets\footnote{\href{https://huggingface.co/collections/locailabs/jupiter-training-data}{\nolinkurl{locailabs/jupiter-training-data}}} are publicly released under open licences.
\end{abstract}

\section{Introduction}

Large Language Models (LLMs) have become foundational infrastructure across industry, government, and research. However, growing dependence on proprietary models concentrated in a small number of jurisdictions poses risks to data sovereignty, national security, and cultural representation~\cite{bondarenko2025sovereign,shi2024culturebank}. Sovereign LLMs, models developed or adapted under national oversight, offer enhanced data protection, safeguard security interests, and contribute to maintaining international competitiveness~\cite{bondarenko2025sovereign}. Yet pre-training frontier models from scratch remains out of reach for the majority of nations: the Llama~3.1 models~\cite{grattafiori2024llama3} required $39.3$ million H100 GPU-hours across all model sizes ($30.84$M for the 405B alone) and produced $11{,}390$ tonnes of CO\textsubscript{2}eq in location-based greenhouse gas emissions~\cite{patterson2022carbon}. This has driven growing interest in developing sovereign capabilities through post-training existing open-weight models, including language transfer~\cite{alexandrov2024bggpt,pipatanakul2026typhoon,alexandrov2024mitigating,joshi2024syntheticlr} and cultural alignment~\cite{shi2024culturebank,xu2025self,masoud2025cultural}. At the same time, LLMs are shifting from passive question-answering toward agentic deployment: autonomous systems that execute multi-step tasks, call tools, and interact with real-world environments~\cite{yang2024swe,wang2026openclaw,merrill2026terminal}. Models serving these use cases must excel at precise instruction following and robust tool use, in addition to any domain-specific capabilities.

We introduce Jupiter-N (where N denotes the Nemotron base), a post-trained variant of Nemotron 3 Super~\cite{nvidia2026nemotron3super}. We select this base because it is, to our knowledge, the strongest fully open model at this parameter count: all weights are released under a permissive licence and all pre-training and post-training datasets are publicly disclosed. Full openness is a deliberate choice for sovereign deployment, where auditability of model behaviour and training provenance is essential for institutional trust and regulatory compliance~\cite{bondarenko2025sovereign}. Additionally, this model is a hybrid reasoner: structured thinking can be enabled or disabled at inference time via the chat template, and Jupiter-N preserves this capability through post-training on both reasoning and non-reasoning traces.

Jupiter-N's post-training targets three objectives:

\begin{enumerate}[topsep=0.5pt, itemsep=-2pt]
  \item \textbf{Agentic.} Improved terminal-use and instruction-following capability to support reliable agentic deployment.
  \item \textbf{Multilingual.} Welsh language support, a language absent from the base model's seven supported languages.
  \item \textbf{Cultural.} UK cultural grounding aligned to British institutional and social norms.
\end{enumerate}

We build on our prior work with Locai L1-Large~\cite{locai2025l1large}, a post-trained Qwen $3$ $235$B model~\cite{yang2025qwen3}, applying the same Forget-Me-Not methodology to mitigate catastrophic forgetting~\cite{mccloskey1989catastrophic,french1999catastrophic}, the tendency for a machine learning model to lose prior capabilities after subsequent fine-tuning. The contributions of this work are threefold:

\begin{enumerate}[label=(\roman*), topsep=0.5pt, itemsep=-2pt]
  \item \textbf{Data curation strategy.} A nine-dataset mixture that balances reasoning and non-reasoning traces to preserve hybrid reasoning, mixes on-policy and off-policy data to mitigate catastrophic forgetting, and introduces an entropy-based curation method for selecting maximally informative training samples (Section~\ref{sec:data}).
  \item \textbf{Training methodology.} A LoRA-based post-training recipe on a LatentMoE architecture with experience replay and role-based loss masking, designed as a replicable template for sovereign post-training (Section~\ref{sec:training}).
  \item \textbf{Comprehensive evaluation.} Benchmarking across mathematical reasoning, instruction following, Welsh-language knowledge, agentic capability, and safety (Section~\ref{sec:eval}).
\end{enumerate}

\section{Base Model}

Nemotron 3 Super~\cite{nvidia2026nemotron3super} (120B parameter, 12B active) employs a Latent Mixture-of-Experts (LatentMoE) architecture, interleaving Mamba-2~\cite{dao2024mamba2} layers, sparse Mixture of Experts (MoE) layers, and Attention layers. The model can support a context window of up to $1$ million tokens and employs Multi-Token Prediction to improve throughput during inference through speculative decoding. We refer the reader to the original technical report~\cite{nvidia2026nemotron3super} for full details on pre-training and post-training.

\section{Data}
\label{sec:data}

\begin{figure}
  \centering
  \vspace{-13pt}
  \includegraphics[width=0.98\columnwidth]{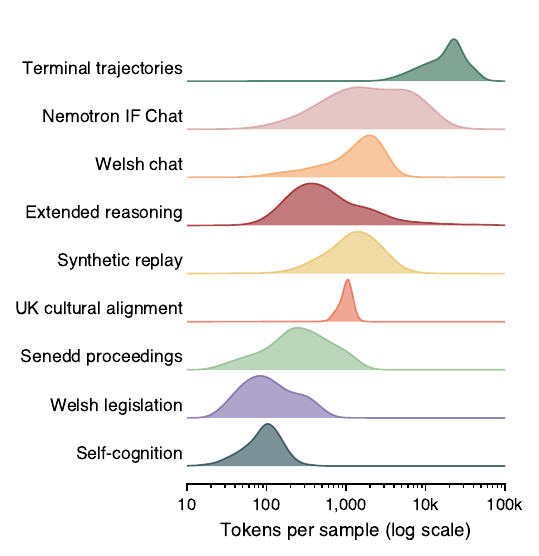}
  \vspace{-0.135cm}
  \caption{Per-dataset sequence length distributions (kernel density estimates on log-scaled token counts). Datasets are ordered by mean length to match Table~\ref{tab:mixture}, spanning three orders of magnitude from self-cognition (${\sim}10^2$) to terminal trajectories (${\sim}10^4$).}
  \label{fig:seq_lengths}
\end{figure}

We curate nine datasets spanning five domains: terminal/agentic capability, cultural alignment, Welsh language, model identity, and general instruction following. All datasets have been open-sourced\footnote{\href{https://huggingface.co/collections/locailabs/jupiter-training-data}{\nolinkurl{locailabs/jupiter-training-data}}}. Table~\ref{tab:mixture} summarises the training mixture and Figure~\ref{fig:seq_lengths} depicts the sequence length distribution per dataset, which span three orders of magnitude, from self-cognition (${\sim}10^2$ tokens) to terminal trajectories (${\sim}10^4$ tokens).

Two design principles guide the mixture. First, as the base model is a hybrid reasoner, we include both reasoning and non-reasoning traces for each applicable dataset to preserve the model's ability to toggle structured thinking at inference time. Second, following our Forget-Me-Not framework~\cite{locai2025l1large}, we mix on-policy data (generated by the unmodified Nemotron 3 Super base) with off-policy data (from external sources and other teacher models) to reduce the distribution shift that leads to catastrophic forgetting~\cite{mccloskey1989catastrophic}. On-policy samples comprise $22$\% of the mixture by sample count. We validated the mixture design and proportion of on-policy data through iterative experiments on the smallest Nemotron 3 variant, Nano 4B, before scaling up to the 120B model. See Appendix~\ref{app:proxy} for details.

\begin{table}[t]
  \centering
  \footnotesize
  \setlength{\tabcolsep}{4pt}
  \begin{tabular}{p{2.8cm}rr}
  \toprule
  \textbf{Dataset} & \textbf{N} & \textbf{Tokens} \\
  \midrule
  Self-cognition & $2$k & $0.2$M \\
  Welsh legislation & $17.9$k & $2.6$M \\
  Senedd proceedings & $19.6$k & $7.2$M \\
  UK cultural alignment & $1.41$k & $1.4$M \\
  Synthetic replay & $8.2$k & $13.3$M \\
  Extended reasoning & $2.06$k & $3.3$M \\
  Welsh chat & $20$k & $35.9$M \\
  Nemotron IF Chat & $15$k & $52.0$M \\
  Terminal trajectories & $30$k & $595.9$M \\
  \bottomrule
  \end{tabular}
  \caption{Training mixture, ordered by mean sequence length. Token counts use the Nemotron tokenizer (131k vocabulary).}
  \label{tab:mixture}
\end{table}

\subsection{Terminal and Agentic Data}
\label{sec:terminal}

The Terminal trajectories dataset is curated from NVIDIA's Nemotron-Terminal-Corpus~\cite{pi2026terminal}, a $366$k-sample Supervised Fine-Tuning (SFT) dataset designed to scale the terminal interaction capabilities of LLMs through multi-step terminal execution trajectories built using their Terminal-Task-Gen pipeline. We select from the \texttt{dataset\_adapters} split ($226$k samples) for its broad coverage across downstream tasks, curated by transforming maths, code, and software engineering datasets into multi-turn terminal trajectories.

\paragraph{Uncertainty-based curation.} Rather than filtering by task outcome, we select samples by information density relative to the target model. Each sample is scored by the unmodified base model: we concatenate the system message and first user message as a prompt, perform greedy decoding for $T = 32$ continuation tokens, and record the top-$k = 20$ log-probabilities at each generated position; restricting to $k{=}20$ avoids materialising the full $131$k-token vocabulary distribution for every position across $226$k samples. We then compute the mean Shannon entropy~\cite{shannon1948mathematical} over these positions. Formally, let $\hat{p}_{t,i}$ denote the renormalised probability of the $i$-th most likely token at position $t$, with $\sum_{i=1}^{k} \hat{p}_{t,i} = 1$. The per-position entropy is
\begin{equation}
  H_t = -\sum_{i=1}^{k} \hat{p}_{t,i} \log \hat{p}_{t,i}
\end{equation}
and the sample-level score is $\bar{H} = \frac{1}{T}\sum_{t=1}^{T} H_t$. As entropy is computed over the renormalised top-$k$ rather than the full vocabulary, absolute values differ from full-distribution entropy; however, the ranking is preserved for selection purposes. High $\bar{H}$ indicates that the model spreads probability across many plausible next tokens, signalling unfamiliarity with the task. We rank all $226$k samples by $\bar{H}$ and retain the top $30{,}000$, those for which the base model is most uncertain. While this criterion is distinct from task success, it is complementary to the original corpus authors' finding that retaining unsuccessful trajectories improves robustness~\cite{pi2026terminal}: hard tasks that the model finds unfamiliar are also more likely to produce failures.

\subsection{UK Cultural Alignment Data}

We generate cultural alignment data using CultureBank~\cite{shi2024culturebank}, a community-driven knowledge base of social norms, values, and everyday practices validated by members of the respective cultural groups. Each training sample combines a user question, persona information describing the questioner's cultural background, and relevant cultural descriptors from CultureBank. We use Qwen $3$ $235$B Instruct~\cite{yang2025qwen3} to generate culturally aware responses in British English, drawing on the cultural knowledge without directly quoting it. We use a non-reasoning model because a thinking trace would leak the cultural context from the system prompt. The system prompt is shown in Figure~\ref{fig:culture_prompt}.

\begin{figure}
\begin{tcolorbox}[colback=blue!5!white, colframe=blue!50!black, fontupper=\small]
You are a British helpful assistant that provides culturally-aware advice about British culture and customs. You have access to relevant cultural judgments that inform your responses. You should provide thoughtful, nuanced advice that considers both the questioner's specific situation and the broader cultural context. Please provide a helpful, culturally-sensitive response to this question. Consider the person's background and situation, and draw insights from the relevant cultural judgments provided above. Your response should be practical, respectful, and demonstrate understanding of British cultural nuances. Do not directly quote the cultural judgments, but use them to inform your response. Do not directly quote the persona, but you can reference the country that the person is from. The user data is the following:\newline
Personal details: \texttt{\{user\_persona\}}\newline
Relevant Cultural Context: \texttt{\{cultural\_context\}}
\end{tcolorbox}
\caption{System prompt used to generate UK cultural alignment training data. Cultural descriptors and persona information are injected at the placeholders.}
\label{fig:culture_prompt}
\end{figure}

\subsection{Self-Cognition Data}

Fine-tuned models frequently hallucinate identities inherited from base weights, prior instruction-tuning stages, or distillation from teacher models. We address this with the self-cognition dataset, a multilingual corpus of identity-related questions covering name, creator, capabilities, limitations, values, and organisational affiliation. From each multi-turn conversation in the source corpus, only the first user message is retained, reducing the task to single-turn question-answering and avoiding conversational dependencies that could conflate identity grounding with dialogue management.

Responses are generated by Nemotron 3 Super with reasoning enabled and with reasoning disabled. A structured system prompt encodes Jupiter-N's identity, Locai Labs' organisational context, and technical provenance, but is excluded from saved training examples so that identity behaviour is learned implicitly rather than conditioned on a runtime system message.

\subsection{Welsh Language Data}

Nemotron 3 Super officially supports seven languages (English, French, German, Italian, Japanese, Spanish, Chinese) but no Celtic languages. The Welsh Government's Cymraeg $2050$ strategy~\cite{welshgov2017cymraeg2050} aims to achieve one million active Welsh speakers by $2050$, and the availability of capable Welsh-language digital tools is recognised as critical infrastructure for that goal. We add Welsh support through two complementary data sources.

\subsubsection{Parallel corpora.} 
We curate two institutional Welsh--English parallel datasets published by Language Technologies Unit, Bangor University\footnote{\href{https://huggingface.co/collections/techiaith/machine-translation-datasets}{\nolinkurl{techiaith/machine-translation-datasets}}}, both produced by professional translators within Welsh public sector institutions: Senedd proceedings ($105$k pairs of parliamentary transcripts) and Welsh legislation ($65$k pairs from legislation.gov.uk), totalling $170$k pairs before deduplication. The two domains are complementary, parliamentary Welsh provides natural, discursive prose covering argumentation and policy discussion, while legal Welsh provides terminologically precise text with consistent formal grammatical constructions. Both are produced by professional translators, giving a substantially lower noise baseline than web-crawled corpora. Each dataset is processed independently through the following stages, preserving source provenance for per-domain ablations.

\paragraph{Cleaning.} Remove pairs where either side has fewer than $20$ characters; remove pairs containing URLs, emoji, or excessive character/word repetition; remove standalone list items. We do not apply source/target length ratio filtering: Welsh morphological inflection and initial consonant mutations produce systematic surface-form differences that inflate character-count ratios on well-formed pairs.

\paragraph{Deduplication.} Deduplication is performed on the English side of each pair: near-duplicate English sources indicate redundant training examples regardless of minor variation in the Welsh translation. We apply a three-stage cascade in order of increasing computational cost so that each stage operates on a strictly smaller set, balancing recall with efficiency: (a) exact deduplication via normalised hash set; (b) near-duplicate detection via MinHash Locality Sensitive Hashing ~\cite{broder1997resemblance} ($1$-gram, $128$ permutations, Jaccard threshold $0.9$), which catches pairs differing only in punctuation or minor word substitutions; (c) semantic deduplication via SemHash~\cite{minishlab2025semhash} (cosine threshold $0.85$), which removes semantically equivalent pairs that differ in surface form.

\paragraph{Instruction formatting.} Clean pairs are converted to chat format: single-turn (${\sim}70$\%) and multi-turn (${\sim}30$\%). Single-turn examples sample uniformly from $21$ instruction templates ($11$ en$\to$cy, $10$ cy$\to$en, including Welsh-medium instructions; see Appendix~\ref{app:templates}), with a balanced 50/50 directional split. Multi-turn examples group $2$--$5$ consecutive pairs from the same document into conversations, exposing the model to sustained translation with shared context.
\subsubsection{Synthetic Welsh chat.} We construct the Welsh chat dataset by translating a portion of NVIDIA's Instruction Following dataset used to train the original Nemotron model\footnote{\href{https://huggingface.co/datasets/nvidia/Nemotron-Instruction-Following-Chat-v1}{\nolinkurl{nvidia/Nemotron-Instruction-Following-Chat-v1}}}. We applied three preprocessing stages: conversation flattening to extract the first user--assistant pair; content filtering to remove self-referential model identity examples ($64{,}157$ rows); and bilingual language detection via \texttt{lingua-language-detector} to discard non-English sources ($12{,}924$ rows), retaining $259{,}750$ rows. Translation uses Qwen $3.5$ $35$B-A$3$B~\cite{yang2025qwen3} in non-thinking mode (using the recommended decoding parameters of temperature $0.7$, top-$p$ $0.8$, top-$k$ $20$, presence penalty $1.5$). Both user and assistant fields are independently translated with the prompt shown in Figure~\ref{fig:translation_prompt}, which preserves XML tags, URLs, mathematical formulas, and code blocks verbatim. $20{,}000$ samples from the translated corpus are included in the training mixture.

\begin{figure}
\begin{tcolorbox}[colback=blue!5!white, colframe=blue!50!black, fontupper=\small]
\textbf{System:} You are an experienced translator that translates documents accurately into fluent text in the target language.\newline\newline
\textbf{User:} You need to translate the following \texttt{\{source\_lang\}} source text to \texttt{\{target\_lang\}}, with the following exceptions:\newline
(1) If the source text contains XML tags, URLs, or math formulas, copy them as-is.\newline
(2) If the source text contains any computer code (e.g.\ C++ or Python), copy the content as-is.\newline\newline
Wrap the translated text in tortoise shell brackets.\newline\newline
\texttt{\{source\_lang\}}: \texttt{\{text\}}
\end{tcolorbox}
\caption{Prompt template for English$\to$Welsh translation of the synthetic Welsh chat dataset. Code, URLs, and mathematical notation are preserved verbatim.}
\label{fig:translation_prompt}
\end{figure}

\subsection{Experience Replay}

Catastrophic forgetting~\cite{mccloskey1989catastrophic,french1999catastrophic} is a well-documented challenge in machine learning in which models lose proficiency on previously learned tasks after subsequent fine-tuning. In the case of LLMs, forgetting can manifest as degraded general capabilities and reasoning, misalignment, or compromised safety~\cite{qi2024finetuning}. Replay, mixing data from the original training distribution into the fine-tuning set, is a simple and effective mitigation~\cite{rolnick2019experience}. Since Nemotron 3 Super is fully open and its post-training data is publicly available, we can combine direct experience replay from the original data with synthetic replay generated by the model itself. Synthetic replay has the additional benefit of being targeted: we can steer generation toward the specific capabilities we wish to preserve, such as chat, instruction following, and reasoning.

Our Forget-Me-Not framework~\cite{locai2025l1large} mitigates forgetting by carefully mixing on-policy replay data with the off-policy task-specific data in the training mixture. We include three replay components, each drawn from the base model's own training data or generated by the unmodified Nemotron 3 Super:

\begin{itemize}[topsep=0.5pt, itemsep=-2pt]
  \item \textbf{Synthetic replay}. We sample Nemotron 3 Super on chat and instruction-following prompts drawn from UltraChat~\cite{ding2023ultrachat}, collecting both reasoning and non-reasoning traces to preserve hybrid reasoning capability. These examples anchor the model's instruction-following and conversational competence during post-training.
  \item \textbf{Extended reasoning traces}\footnote{\href{https://huggingface.co/datasets/RamAnanth1/Nemotron3-Super-Reasoning-2000x}{\nolinkurl{Nemotron3-Super-Reasoning-2000x}}}. Reasoning-mode outputs generated by Nemotron 3 Super, included to retain the model's chain-of-thought reasoning capability.
  \item \textbf{Nemotron IF Chat}\footnote{\href{https://huggingface.co/datasets/nvidia/Nemotron-Instruction-Following-Chat-v1}{\nolinkurl{nvidia/Nemotron-Instruction-Following-Chat-v1}}}. A subset of the base model's original post-training data, providing direct experience replay that reinforces instruction-following behaviour.
\end{itemize}

\begin{figure}[b]
  \centering
  \vspace{-10pt}
  \includegraphics[width=\columnwidth]{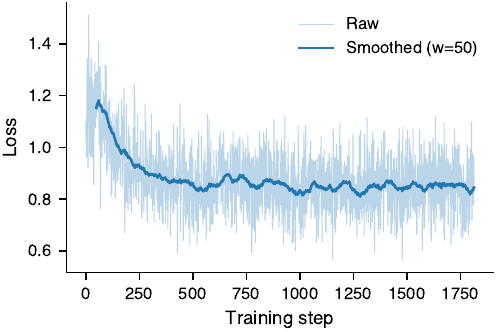}
  \vspace{-17pt}
  \caption{Training loss over the single-epoch fine-tuning run.}
  \label{fig:loss_curve}
\end{figure}

\begin{table*}[t]
  \centering
  \footnotesize
  \setlength{\tabcolsep}{4pt}
  \begin{tabular}{ll rr rr}
  \toprule
  & & \multicolumn{2}{c}{\textit{Reasoning off}} & \multicolumn{2}{c}{\textit{Reasoning on}} \\
  \cmidrule(lr){3-4} \cmidrule(lr){5-6}
  \textbf{Benchmark} & \textbf{Metric} & \textbf{Jupiter-N} & \textbf{Nemotron} & \textbf{Jupiter-N} & \textbf{Nemotron} \\
  \midrule
  IFEval & prompt strict & $\mathbf{80.96}$ & $79.85$ & $\mathbf{90.20}$ & $\mathbf{90.20}$ \\
  IFBench & prompt loose & $\mathbf{41.8}$ & $37.4$ & $\mathbf{73.8}$ & $69.7$ \\
  AgentHarm & harm $\downarrow$ & $\mathbf{73.4}$ & $78.6$ & $\mathbf{53.8}$ & $55.4$ \\
  Terminal Bench 2 & accuracy & $-$ & $-$ & $\mathbf{52.7}$ & $43.6$ \\
  GSM8K & accuracy & $-$ & $-$ & $\mathbf{94.01}$ & $93.56$ \\
  Welsh ARC-Easy & accuracy & $\mathbf{72.00}$ & $54.00$ & $-$ & $-$ \\
  Welsh MMLU-Lite & accuracy & $\mathbf{61.25}$ & $56.00$ & $-$ & $-$ \\
  \bottomrule
  \end{tabular}
  \caption{Evaluation results (all values \%). Nemotron denotes the unmodified Nemotron 3 Super base. Best result per benchmark for each reasoning mode is \textbf{bolded}. Both models use temperature $1.0$, top-$p$ $0.95$.}
  \label{tab:results}
\end{table*}

\section{Training}
\label{sec:training}

All nine datasets are merged into a single shuffled corpus and used for $1$-epoch Low Rank Adaptation (LoRA)~\cite{hu2022lora} training. We use rank $16$, alpha $32$, with Mamba \texttt{out\_proj} layers excluded (these use custom kernels incompatible with LoRA). Training uses FSDP2 with expert parallelism across $8$ H200 GPUs, activation checkpointing, a global batch size of $64$ (local $8$), sequence length $2{,}048$, Adam optimiser ($\beta_1{=}0.9$, $\beta_2{=}0.999$), and cosine learning rate decay from $1 \times 10^{-5}$ to $1 \times 10^{-6}$. We apply role-based loss masking in which the cross-entropy loss is computed only over assistant-role tokens, with system and user turns masked to zero weight. This ensures gradients flow exclusively from response generation, preventing the model from wasting capacity learning to predict prompts or system instructions. Figure~\ref{fig:loss_curve} shows the training loss over the single epoch.

\section{Evaluation}
\label{sec:eval}

\subsection{Benchmarks}

We evaluate Jupiter-N and Nemotron 3 Super across the following benchmarks. IFEval and GSM8K are evaluated using LightEval\footnote{\url{https://github.com/huggingface/lighteval}}; all other benchmarks use their official evaluation repositories.

\paragraph{Instruction following.} IFEval~\cite{zhou2023ifeval} contains $541$ prompts with $25$ verifiable constraint types (e.g.\ word count, format, keyword inclusion); we report prompt-level strict accuracy, the standard metric used by the Open LLM Leaderboard and most prior work. IFBench~\cite{pyatkin2025ifbench} extends this to compositional constraints, requiring models to satisfy multiple simultaneous requirements within a single response; we report prompt-level loose accuracy, the primary metric recommended by the authors. To reflect real world usage, we evaluate in both reasoning and non-reasoning mode.

\paragraph{Mathematical reasoning.} GSM8K~\cite{cobbe2021gsm8k} contains $1{,}319$ grade-school maths word problems requiring multi-step arithmetic reasoning. Following the Nemotron evaluation setup for mathematical benchmarks~\cite{nvidia2026nemotron3super}, we report accuracy with reasoning enabled.

\paragraph{Agentic and terminal.} Terminal Bench 2~\cite{merrill2026terminal} evaluates multi-step terminal task execution, requiring models to plan, issue, and chain shell commands to accomplish file-system, package-management, and system-administration objectives. We report accuracy on the medium-difficulty subset ($55$ tasks). Following the original Nemotron evaluation~\cite{nvidia2026nemotron3super}, reasoning is enabled for agentic evaluation.

\paragraph{Safety.} AgentHarm~\cite{andriushchenko2025agentharm} presents $110$ malicious agent tasks spanning $11$ harm categories (fraud, cyberattack, disinformation, etc.), developed by the UK AI Safety Institute. We report the harmful task completion rate (lower is better). To reflect real world usage we evaluate in both reasoning and non-reasoning mode.

\paragraph{Welsh language.} Welsh MMLU-Lite ($400$ questions) and Welsh ARC-Easy ($50$ questions) from Bangor University's \texttt{llm-evals-cy}~\cite{techiaith2026llmevals} are Welsh-language adaptations of their English counterparts, testing factual knowledge and elementary science reasoning in Welsh. We evaluate with reasoning disabled to isolate language proficiency from chain-of-thought effects.

\subsection{Inference}

We serve both Jupiter-N and Nemotron 3 Super with vLLM using tensor parallelism and expert parallelism across $4$ GPUs. The KV cache is stored in FP8 and the Mamba SSM state cache in float16. All evaluations use nucleus sampling with temperature $1.0$ and top-$p$ $0.95$. 

\subsection{Results}

Table~\ref{tab:results} compares Jupiter-N against the unmodified Nemotron 3 Super base. The clearest gains appear in the three domains we explicitly target. Welsh fluency improves greatly with $+18.00$ on ARC-Easy and $+5.25$ on MMLU-Lite, attributable to the parallel-corpus and synthetic-chat pipeline described in Section~\ref{sec:data}. Instruction following improves across both reasoning modes, with IFBench rising by $+4.4$ (reasoning off) and $+4.1$ (reasoning on), and IFEval by $+1.11$ without reasoning while matching the base with reasoning. On Terminal Bench~2, Jupiter-N outperforms the base by $+9.1$ points, demonstrating the benefit of the entropy-based curation strategy that prioritises trajectories the base model finds most unfamiliar (Section~\ref{sec:terminal}).

Critically, these gains do not come at the expense of existing capabilities. GSM8K is retained at near-parity ($94.01$ vs $93.56$), indicating that the Forget-Me-Not replay strategy successfully mitigates catastrophic forgetting. Safety also improves, with AgentHarm harmful rates decreasing by $-5.2$ (reasoning off) and $-1.6$ (reasoning on).

\section{Conclusion}

We have presented Jupiter-N, a post-trained variant of Nemotron 3 Super that adds Welsh language support, improves agentic and instruction-following capability, and introduces UK cultural grounding. The key insight is that careful mixture design, combining on-policy experience replay with off-policy task data via our Forget-Me-Not framework, enables targeted capability injection while preserving the base model's existing strengths. We additionally introduce an entropy-based curation strategy that selects training samples the base model finds most unfamiliar, improving sample efficiency for agentic training data. Evaluation confirms gains across all targeted domains with no meaningful regression on mathematical reasoning or safety. We frame this work as a reproducible template for \emph{sovereign post-training}: substituting cultural knowledge bases, institutional corpora, and target languages produces an equivalent pipeline for any country.

\section{Limitations}

Welsh evaluation relies on Welsh-adapted versions of English-origin benchmarks (ARC-Easy, MMLU), which test factual recall in Welsh but do not assess native Welsh natural language understanding tasks. The benchmarks are also small ($50$ questions for ARC-Easy, $400$ for MMLU-Lite), so individual results may exhibit high variance. The Welsh parallel corpora are drawn exclusively from formal institutional domains (parliamentary and legal), so the resulting model may underperform on colloquial or informal Welsh, and model outputs in Welsh have not yet undergone extensive human quality review. Additionally, the cultural grounding introduced via CultureBank-informed data has not been validated through human evaluation. Finally, the self-cognition data is generated by teacher models and may not generalise to adversarial identity probing beyond the templates used.

\section{Environmental Impact}

Post-training completed in $9$ hours $15$ minutes on $8$ H200 GPUs powered by $100$\% renewable energy. This consumed an estimated $52$~kWh of GPU energy ($8 \times 700$\,W TDP $\times$ $9.25$\,h), or ${\sim}57$~kWh after applying a power usage effectiveness (PUE) of $1.1$. As the data centre runs entirely on renewable electricity, the location-based operational carbon footprint is effectively zero~\cite{patterson2022carbon}. We report these figures following recommendations for transparent energy accounting in machine learning research.

\bibliography{custom}

@misc{locai2025l1large,
  title={Locai {L1-Large}: An Open-Source Instruct Model},
  author={Drayson, George},
  year={2025},
  howpublished={\url{https://locailabs.com/blog/technical-blog}},
  note={Blog post}
}

@article{french1999catastrophic,
  title={Catastrophic forgetting in connectionist networks},
  author={French, Robert M},
  journal={Trends in cognitive sciences},
  volume={3},
  number={4},
  pages={128--135},
  year={1999},
  publisher={Elsevier}
}

@incollection{mccloskey1989catastrophic,
  title={Catastrophic interference in connectionist networks: The sequential learning problem},
  author={McCloskey, Michael and Cohen, Neal J},
  booktitle={Psychology of learning and motivation},
  volume={24},
  pages={109--165},
  year={1989},
  publisher={Elsevier}
}

@article{rolnick2019experience,
  title={Experience replay for continual learning},
  author={Rolnick, David and Ahuja, Arun and Schwarz, Jonathan and Lillicrap, Timothy and Wayne, Gregory},
  journal={Advances in neural information processing systems},
  volume={32},
  year={2019}
}

@inproceedings{andriushchenko2025agentharm,
title={AgentHarm: A Benchmark for Measuring Harmfulness of {LLM} Agents},
author={Maksym Andriushchenko and Alexandra Souly and Mateusz Dziemian and Derek Duenas and Maxwell Lin and Justin Wang and Dan Hendrycks and Andy Zou and J Zico Kolter and Matt Fredrikson and Yarin Gal and Xander Davies},
booktitle={The Thirteenth International Conference on Learning Representations},
year={2025},
url={https://openreview.net/forum?id=AC5n7xHuR1}
}

@misc{minishlab2025semhash,
  author       = {{van Dongen}, Thomas and Stephan Tulkens},
  title        = {SemHash: Fast Multimodal Semantic Deduplication \& Filtering},
  year         = {2025},
  publisher    = {Zenodo},
  doi          = {10.5281/zenodo.17265942},
  url          = {https://github.com/MinishLab/semhash},
  license      = {MIT}
}

@inproceedings{ding2023ultrachat,
    title = "Enhancing Chat Language Models by Scaling High-quality Instructional Conversations",
    author = "Ding, Ning  and
      Chen, Yulin  and
      Xu, Bokai  and
      Qin, Yujia  and
      Hu, Shengding  and
      Liu, Zhiyuan  and
      Sun, Maosong  and
      Zhou, Bowen",
    editor = "Bouamor, Houda  and
      Pino, Juan  and
      Bali, Kalika",
    booktitle = "Proceedings of the 2023 Conference on Empirical Methods in Natural Language Processing",
    month = dec,
    year = "2023",
    address = "Singapore",
    publisher = "Association for Computational Linguistics",
    url = "https://aclanthology.org/2023.emnlp-main.183/",
    doi = "10.18653/v1/2023.emnlp-main.183",
    pages = "3029--3051",
}

@inproceedings{shi2024culturebank,
    title = "{C}ulture{B}ank: An Online Community-Driven Knowledge Base Towards Culturally Aware Language Technologies",
    author = "Shi, Weiyan  and
      Li, Ryan  and
      Zhang, Yutong  and
      Ziems, Caleb  and
      Yu, Sunny  and
      Horesh, Raya  and
      Paula, Rog{\'e}rio Abreu De  and
      Yang, Diyi",
    editor = "Al-Onaizan, Yaser  and
      Bansal, Mohit  and
      Chen, Yun-Nung",
    booktitle = "Findings of the Association for Computational Linguistics: EMNLP 2024",
    month = nov,
    year = "2024",
    address = "Miami, Florida, USA",
    publisher = "Association for Computational Linguistics",
    url = "https://aclanthology.org/2024.findings-emnlp.288/",
    doi = "10.18653/v1/2024.findings-emnlp.288",
    pages = "4996--5025",
}

@inproceedings{pyatkin2025ifbench,
title={Generalizing Verifiable Instruction Following},
author={Valentina Pyatkin and Saumya Malik and Victoria Graf and Hamish Ivison and Shengyi Huang and Pradeep Dasigi and Nathan Lambert and Hannaneh Hajishirzi},
booktitle={The Thirty-ninth Annual Conference on Neural Information Processing Systems Datasets and Benchmarks Track},
year={2025},
url={https://openreview.net/forum?id=yfYgwjj5F8}
}

@misc{nvidia2026nemotron3super,
  title  = {NVIDIA Nemotron 3: Efficient and Open Intelligence},
  author = {{NVIDIA}},
  year   = {2025},
  url    = {https://arxiv.org/abs/2512.20856},
  note   = {White Paper}
}

@misc{techiaith2026llmevals,
  title={llm-evals-cy: Welsh Language Evaluation Suite for Large Language Models},
  author={{Techiaith, Bangor University}},
  year={2026},
  howpublished={\url{https://github.com/techiaith/llm-evals-cy}}
}

@inproceedings{broder1997resemblance,
  title={On the resemblance and containment of documents},
  author={Broder, Andrei Z},
  booktitle={Proceedings. Compression and Complexity of SEQUENCES 1997 (Cat. No. 97TB100171)},
  pages={21--29},
  year={1997},
  organization={IEEE}
}

@article{joshi2024syntheticlr,
  title={Adapting Multilingual {LLMs} to Low-Resource Languages using Continued Pre-training and Synthetic Corpus},
  author={Joshi, Raviraj and Singla, Kanishk and Kamath, Anusha and Kalani, Raunak and Paul, Rakesh and Vaidya, Utkarsh and Chauhan, Sanjay Singh and Wartikar, Niranjan and Long, Eileen},
  journal={arXiv preprint arXiv:2410.14815},
  year={2024}
}

@inproceedings{hu2022lora,
title={Lo{RA}: Low-Rank Adaptation of Large Language Models},
author={Edward J Hu and yelong shen and Phillip Wallis and Zeyuan Allen-Zhu and Yuanzhi Li and Shean Wang and Lu Wang and Weizhu Chen},
booktitle={International Conference on Learning Representations},
year={2022},
url={https://openreview.net/forum?id=nZeVKeeFYf9}
}

@article{bondarenko2025sovereign,
  title={Sovereign Large Language Models: Advantages, Strategy and Regulations},
  author={Bondarenko, Mykhailo and Lushnei, Sviatoslav and Paniv, Yurii and Molchanovsky, Oleksii and Romanyshyn, Mariana and Filipchuk, Yurii and Kiulian, Artur},
  journal={arXiv preprint arXiv:2503.04745},
  year={2025}
}

@article{alexandrov2024bggpt,
  title={{BgGPT} 1.0: Extending English-centric {LLMs} to Other Languages},
  author={Alexandrov, Anton and Raychev, Veselin and Dimitrov, Dimitar I and Zhang, Ce and Vechev, Martin and Toutanova, Kristina},
  journal={arXiv preprint arXiv:2412.10893},
  year={2024}
}

@inproceedings{alexandrov2024mitigating,
    title = "Mitigating Catastrophic Forgetting in Language Transfer via Model Merging",
    author = {Alexandrov, Anton  and
      Raychev, Veselin  and
      M{\"u}ller, Mark Niklas  and
      Zhang, Ce  and
      Vechev, Martin  and
      Toutanova, Kristina},
    editor = "Al-Onaizan, Yaser  and
      Bansal, Mohit  and
      Chen, Yun-Nung",
    booktitle = "Findings of the Association for Computational Linguistics: EMNLP 2024",
    month = nov,
    year = "2024",
    address = "Miami, Florida, USA",
    publisher = "Association for Computational Linguistics",
    url = "https://aclanthology.org/2024.findings-emnlp.1000/",
    doi = "10.18653/v1/2024.findings-emnlp.1000",
    pages = "17167--17186",
}

@article{yang2024swe,
  title={{SWE}-agent: Agent-computer interfaces enable automated software engineering},
  author={Yang, John and Jimenez, Carlos E and Wettig, Alexander and Lieret, Kilian and Yao, Shunyu and Narasimhan, Karthik and Press, Ofir},
  journal={Advances in Neural Information Processing Systems},
  volume={37},
  pages={50528--50652},
  year={2024}
}

@inproceedings{dao2024mamba2,
title={Transformers are {SSM}s: Generalized Models and Efficient Algorithms Through Structured State Space Duality},
author={Tri Dao and Albert Gu},
booktitle={Forty-first International Conference on Machine Learning},
year={2024},
url={https://openreview.net/forum?id=ztn8FCR1td}
}

@inproceedings{
merrill2026terminal,
title={Terminal-Bench: Benchmarking Agents on Hard, Realistic Tasks in Command Line Interfaces},
  author={Merrill, Mike A and Shaw, Alexander G and Carlini, Nicholas and Li, Boxuan and Raj, Harsh and Bercovich, Ivan and Shi, Lin and Shin, Jeong Yeon and Walshe, Thomas and Buchanan, E Kelly and others},
booktitle={The Fourteenth International Conference on Learning Representations},
year={2026},
url={https://openreview.net/forum?id=a7Qa4CcHak}
}

@article{zhou2023ifeval,
  title={Instruction-following evaluation for large language models},
  author={Zhou, Jeffrey and Lu, Tianjian and Mishra, Swaroop and Brahma, Siddhartha and Basu, Sujoy and Luan, Yi and Zhou, Denny and Hou, Le},
  journal={arXiv preprint arXiv:2311.07911},
  year={2023}
}

@article{wang2026openclaw,
  title={{OpenClaw-RL}: Train any agent simply by talking},
  author={Wang, Yinjie and Chen, Xuyang and Jin, Xiaolong and Wang, Mengdi and Yang, Ling},
  journal={arXiv preprint arXiv:2603.10165},
  year={2026}
}

@misc{pi2026terminal,
      title={On Data Engineering for Scaling LLM Terminal Capabilities}, 
      author={Renjie Pi and Grace Lam and Mohammad Shoeybi and Pooya Jannaty and Bryan Catanzaro and Wei Ping},
      year={2026},
      eprint={2602.21193},
      archivePrefix={arXiv},
      primaryClass={cs.CL},
      url={https://arxiv.org/abs/2602.21193}, 
}

@article{shannon1948mathematical,
  title={A mathematical theory of communication},
  author={Shannon, Claude Elwood},
  journal={The Bell system technical journal},
  volume={27},
  number={3},
  pages={379--423},
  year={1948},
  publisher={Nokia Bell Labs}
}

@article{yang2025qwen3,
  title={Qwen3 technical report},
  author={Yang, An and Li, Anfeng and Yang, Baosong and Zhang, Beichen and Hui, Binyuan and Zheng, Bo and Yu, Bowen and Gao, Chang and Huang, Chengen and Lv, Chenxu and others},
  journal={arXiv preprint arXiv:2505.09388},
  year={2025}
}

@article{grattafiori2024llama3,
  title={The llama 3 herd of models},
  author={Grattafiori, Aaron and Dubey, Abhimanyu and Jauhri, Abhinav and Pandey, Abhinav and Kadian, Abhishek and Al-Dahle, Ahmad and Letman, Aiesha and Mathur, Akhil and Schelten, Alan and Vaughan, Alex and others},
  journal={arXiv preprint arXiv:2407.21783},
  year={2024}
}

@article{pipatanakul2026typhoon,
  title={Typhoon-S: Minimal Open Post-Training for Sovereign Large Language Models},
  author={Pipatanakul, Kunat and Taveekitworachai, Pittawat},
  journal={arXiv preprint arXiv:2601.18129},
  year={2026}
}

@inproceedings{xu2025self,
    title = "Self-Pluralising Culture Alignment for Large Language Models",
    author = "Xu, Shaoyang  and
      Leng, Yongqi  and
      Yu, Linhao  and
      Xiong, Deyi",
    editor = "Chiruzzo, Luis  and
      Ritter, Alan  and
      Wang, Lu",
    booktitle = "Proceedings of the 2025 Conference of the Nations of the Americas Chapter of the Association for Computational Linguistics: Human Language Technologies (Volume 1: Long Papers)",
    month = apr,
    year = "2025",
    address = "Albuquerque, New Mexico",
    publisher = "Association for Computational Linguistics",
    url = "https://aclanthology.org/2025.naacl-long.350/",
    doi = "10.18653/v1/2025.naacl-long.350",
    pages = "6859--6877",
    ISBN = "979-8-89176-189-6",
}

@inproceedings{masoud2025cultural,
    title = "Cultural Alignment in Large Language Models: An Explanatory Analysis Based on Hofstede{'}s Cultural Dimensions",
    author = "Masoud, Reem I.  and
      Liu, Ziquan  and
      Ferianc, Martin  and
      Treleaven, Philip  and
      Rodrigues, Miguel",
    editor = "Rambow, Owen  and
      Wanner, Leo  and
      Apidianaki, Marianna  and
      Al-Khalifa, Hend  and
      Eugenio, Barbara Di  and
      Schockaert, Steven",
    booktitle = "Proceedings of the 31st International Conference on Computational Linguistics",
    month = jan,
    year = "2025",
    address = "Abu Dhabi, UAE",
    publisher = "Association for Computational Linguistics",
    url = "https://aclanthology.org/2025.coling-main.567/",
    pages = "8474--8503",
}

@inproceedings{
qi2024finetuning,
title={Fine-tuning Aligned Language Models Compromises Safety, Even When Users Do Not Intend To!},
author={Xiangyu Qi and Yi Zeng and Tinghao Xie and Pin-Yu Chen and Ruoxi Jia and Prateek Mittal and Peter Henderson},
booktitle={The Twelfth International Conference on Learning Representations},
year={2024},
url={https://openreview.net/forum?id=hTEGyKf0dZ}
}

@article{cobbe2021gsm8k,
  title={Training verifiers to solve math word problems},
  author={Cobbe, Karl and Kosaraju, Vineet and Bavarian, Mohammad and Chen, Mark and Jun, Heewoo and Kaiser, Lukasz and Plappert, Matthias and Tworek, Jerry and Hilton, Jacob and Nakano, Reiichiro and others},
  journal={arXiv preprint arXiv:2110.14168},
  year={2021}
}

@article{patterson2022carbon,
  title={The carbon footprint of machine learning training will plateau, then shrink},
  author={Patterson, David and Gonzalez, Joseph and H{\"o}lzle, Urs and Le, Quoc and Liang, Chen and Munguia, Lluis-Miquel and Rothchild, Daniel and So, David R and Texier, Maud and Dean, Jeff},
  journal={Computer},
  volume={55},
  number={7},
  pages={18--28},
  year={2022},
  publisher={IEEE}
}

@misc{welshgov2017cymraeg2050,
  title={Cymraeg 2050: A Million Welsh Speakers},
  author={{Welsh Government}},
  year={2017},
  howpublished={\url{https://www.gov.wales/cymraeg-2050-welsh-language-strategy}}
}

\appendix

\section{Proxy Experiments on Nemotron 4B}
\label{app:proxy}

Before committing to the full $120$B training run, we used Nemotron 3 Nano 4B as a rapid prototyping proxy to iteratively refine the data mixture. All runs include self-cognition and UK cultural alignment data as a fixed baseline; the tables show only the components that vary between iterations. Hyperparameters (learning rate, weight decay, batch size, epochs) were co-tuned between iterations, so individual rows should not be interpreted as single-variable ablations; the tables reflect the development trajectory rather than a controlled experiment. We report two sets of experiments evaluated under different reasoning modes.

\subsection{Effect of Synthetic Replay} Table~\ref{tab:proxy_replay} isolates the impact of adding synthetic replay to the task data. Synthetic replay partially recovered IFEval from $86.1$ to $88.2$, close to but not fully matching the base model's $88.7$. The residual gap motivated two decisions for the final $120$B mixture: including direct experience replay from the base model's original training data, and adding on-policy instruction-following data (Nemotron IF Chat). Together, these not only recovered but improved instruction following at $120$B scale (Table~\ref{tab:results}).

\begin{table}[h]
\centering
\footnotesize
\setlength{\tabcolsep}{5pt}
\begin{tabular}{lrr}
\toprule
\textbf{Configuration} & \textbf{IF} & \textbf{Maths} \\
\midrule
Base (Nemotron 4B) & $\mathbf{88.7}$ & $\mathbf{85.8}$ \\
No synthetic replay & $86.1$ & $81.7$ \\
+ synthetic replay & $88.2$ & $81.1$ \\
\bottomrule
\end{tabular}
\caption{Effect of synthetic replay on Nemotron 3 Nano 4B. IF = IFEval prompt strict, Maths = GSM8K (all values \%, reasoning enabled).}
\label{tab:proxy_replay}
\end{table}

\subsection{Effect of Welsh Data} Table~\ref{tab:proxy_welsh} tracks the addition of Welsh data to the replay mixture, evaluated with reasoning disabled. Substituting parallel corpora for synthetic Welsh chat dramatically improved Welsh MMLU-Lite ($+8.25$) while maintaining GSM8K, justifying the synthetic data strategy. The IFEval regression at $4$B scale ($-6.8$ with Welsh synthetic) underscored the importance of sufficient replay and on-policy instruction-following data when adding new task domains; in the final $120$B mixture, the combination of direct experience replay and Nemotron IF Chat not only recovers but improves instruction following over the base (Table~\ref{tab:results}).

\begin{table}[h]
\centering
\footnotesize
\setlength{\tabcolsep}{3pt}
\begin{tabular}{p{2.8cm}rrr}
\toprule
\textbf{Configuration} & \textbf{IF} & \textbf{Maths} & \textbf{Welsh} \\
\midrule
Base (Nemotron 4B) & $\mathbf{83.7}$ & $84.1$ & $29.25$ \\
Synthetic replay & $80.0$ & $85.0$ & $--$ \\
+ Welsh parallel & $79.9$ & $84.2$ & $30.75$ \\
+ Welsh synthetic & $76.9$ & $\mathbf{85.4}$ & $\mathbf{37.50}$ \\
\bottomrule
\end{tabular}
\caption{Effect of Welsh data on Nemotron 3 Nano 4B. IF = IFEval prompt strict, Maths = GSM8K, Welsh = Welsh MMLU-Lite (all values \%, reasoning disabled).}
\label{tab:proxy_welsh}
\end{table}

\section{Welsh Translation Instruction Templates}
\label{app:templates}

Tables~\ref{tab:en2cy} and~\ref{tab:cy2en} list the instruction templates used to format Welsh parallel corpora into single-turn chat examples (Section~\ref{sec:data}). Each pair is assigned a template sampled uniformly at random, with a balanced 50/50 directional split.

\begin{table*}[h]
\centering
\small
\begin{tabular}{rl}
\toprule
\# & \textbf{English$\to$Welsh template} \\
\midrule
1 & Translate the following English text into Welsh: \texttt{\{text\}} \\
2 & Translate to Welsh: \texttt{\{text\}} \\
3 & What is the Welsh for the following? \texttt{\{text\}} \\
4 & Please translate this into Welsh: \texttt{\{text\}} \\
5 & Convert the following to Welsh: \texttt{\{text\}} \\
6 & Provide a Welsh translation of: \texttt{\{text\}} \\
7 & How would you say this in Welsh? \texttt{\{text\}} \\
8 & Render the following in Welsh: \texttt{\{text\}} \\
9 & Cyfieithwch y testun Saesneg canlynol i'r Gymraeg: \texttt{\{text\}} \\
10 & Cyfieithwch i'r Gymraeg: \texttt{\{text\}} \\
11 & Beth yw'r cyfieithiad Cymraeg o'r canlynol? \texttt{\{text\}} \\
\bottomrule
\end{tabular}
\caption{English$\to$Welsh instruction templates ($n=11$).}
\label{tab:en2cy}
\end{table*}

\begin{table*}[h]
\centering
\small
\begin{tabular}{rl}
\toprule
\# & \textbf{Welsh$\to$English template} \\
\midrule
1 & Translate the following Welsh text into English: \texttt{\{text\}} \\
2 & Translate to English: \texttt{\{text\}} \\
3 & What is the English for the following? \texttt{\{text\}} \\
4 & Please translate this Welsh text into English: \texttt{\{text\}} \\
5 & Convert the following Welsh to English: \texttt{\{text\}} \\
6 & Provide an English translation of this Welsh text: \texttt{\{text\}} \\
7 & How would you say this in English? \texttt{\{text\}} \\
8 & Render the following Welsh text in English: \texttt{\{text\}} \\
9 & Cyfieithwch y testun Cymraeg canlynol i'r Saesneg: \texttt{\{text\}} \\
10 & Rhowch gyfieithiad Saesneg o'r testun Cymraeg canlynol: \texttt{\{text\}} \\
\bottomrule
\end{tabular}
\caption{Welsh$\to$English instruction templates ($n=10$).}
\label{tab:cy2en}
\end{table*}

\end{document}